\documentclass[runningheads]{llncs}
\usepackage{booktabs}
\usepackage{graphicx}
\usepackage{multirow}
\usepackage{amssymb}
\usepackage{pythonhighlight}
\usepackage{amsmath}
\usepackage{wrapfig}
\usepackage{bbding}
\definecolor{ao}{rgb}{0.01, 0.75, 0.24}
\usepackage[colorlinks,linkcolor=red, anchorcolor=blue, citecolor=blue, urlcolor=magenta]{hyperref}

\begin{document}
%
\title{Flexible Sampling for Long-tailed Skin Lesion Classification}
\titlerunning{Flexible Sampling}

\author{Lie Ju\inst{1,2}, 
Yicheng Wu\inst{3}, 
Lin Wang\inst{1, 4}, 
Zhen Yu\inst{1,2},
Xin Zhao\inst{1},  \\
Xin Wang\inst{1}, 
Paul Bonnington\inst{2} and 
Zongyuan Ge\inst{1,2(}\Envelope\inst{)}}

\authorrunning{L. Ju et al.}
\institute{Monash-Airdoc Research, Monash University, Melbourne, Australia \and
Monash Medical AI Group, Monash University, Melbourne, Australia \and
Faculty of Information Technology, Monash University, Melbourne, Australia \and
Harbin Engineering University, Harbin, China
\\ \url{https://mmai.group}
\\
\email{julie@airdoc.com}, \email{zongyuan.ge@monash.edu}}

\maketitle              
\begin{abstract}
Most of the medical tasks naturally exhibit a long-tailed distribution due to the complex patient-level conditions and the existence of rare diseases. Existing long-tailed learning methods usually treat each class equally to re-balance the long-tailed distribution. However, considering that some challenging classes may present diverse intra-class distributions, re-balancing all classes equally may lead to a significant performance drop. 
To address this, in this paper, we propose a curriculum learning-based framework called Flexible Sampling for the long-tailed skin lesion classification task. 
Specifically, we initially sample a subset of training data as anchor points based on the individual class prototypes. Then, these anchor points are used to pre-train an inference model to evaluate the per-class learning difficulty. Finally, we use a curriculum sampling module to dynamically query new samples from the rest training samples with the learning difficulty-aware sampling probability. We evaluated our model against several state-of-the-art methods on the ISIC dataset. The results with two long-tailed settings have demonstrated the superiority of our proposed training strategy, which achieves a new benchmark for long-tailed skin lesion classification.
\keywords{Long-tailed classification \and Skin lesion \and Flexible Sampling}
\end{abstract}

\section{Introduction}
In the real world, since the conditions of patients are complex and there is a wide range of disease categories~\cite{england2016diagnostic}, most of the medical datasets tend to exhibit a long-tailed distribution characteristics~\cite{ju2021relational}. For instance, in dermatology, skin lesions can be divided into different subtypes according to their particular clinical characteristics~\cite{esteva2017dermatologist}. On the popular ISIC dataset~\cite{isic}, the ratio of majority classes and minority classes exceeds one hundred, which presents a highly-imbalanced class distribution. Deep classification models trained on such a long-tailed distribution might not recognize those minority classes well, since there are only a few samples available~\cite{liu2019large}. Improving the recognition accuracy of those rare diseases is highly desirable for the practical applications.

Most of existing methods for long-tailed learning can be roughly divided into three main categories:
information augmentation, module improvement~\cite{zhang2021deep} and class-balancing. The information augmentation-based~\cite{zhang2017mixup,liu2019large} methods focused on building a shared memory between head and tail but failed to tackle extremely imbalanced conditions since there are still more head samples for a mini-batch sampling. The module improvement-based methods~\cite{wang2020long,ju2021relational} required multiple experts for the final decision and introduces higher computational costs. The class-balancing methods~\cite{kang2019decoupling,zhou2020bbn} attempted to re-balance the class distribution, which becomes the mainstream model for various long-tailed classification tasks. However, most existing methods ignored or underestimated the differences of the class-level learning difficulties. In other words, we believe that it is crucial to dynamically handle different classes according to their learning difficulties.
\begin{figure}[t]
    \centering
    \includegraphics[width=10cm]{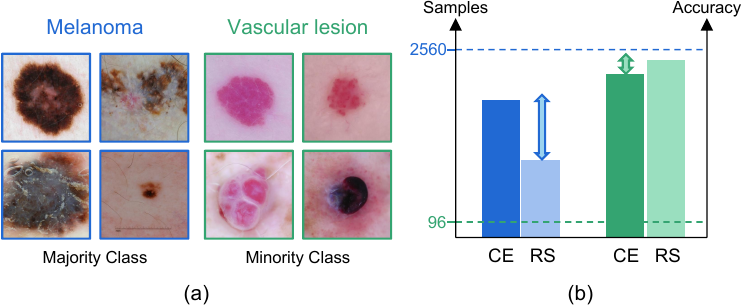}
    \caption{An illustration of \textbf{learning difficulty}. RS denotes re-sampling. The sub-figure (a) shows two kinds of lesions melanoma and vascular lesion from the majority and minority classes respectively. The sub-figure (b) shows that naively over-sampling those minority classes can only achieve marginal performance gains on VASC but seriously hurts the recognition accuracy on MEL.}
    \label{fig.problem}
\end{figure}

In Fig.~\ref{fig.problem}, we illustrate our motivation with a pilot experiment on ISIC. \textit{Melanoma (MEL)} and \textit{vascular lesion (VASC)} are two representative skin lesions from the majority and minority classes, respectively. The former class has 2560 images for training while the latter only contains 96 training samples. 
It can be seen that most samples from VASC resemble one another and present a pink color. Intuitively, this kind of lesions is easier to be classified by deep models since the intra-class diversity is relatively small. By contrast, although MEL has more samples for training, various shapes and textures make the learning difficulty higher. Fig.~\ref{fig.problem}(b) shows the performance of two models trained with different loss functions (cross-entropy (CE) loss vs. re-sampling (RS)) for the class-balancing constraint. It can be seen that over-sampling the tailed classes with less intra-class variety leads to marginal performance gains on VASC but a significant performance drop on MEL, also known as a severe marginal effect.

From this perspective, we propose a novel curriculum learning-based framework named Flexible Sampling through considering the varied learning difficulties of different classes. Specifically, first, we initially sample a subset of data from the original distribution as the \textit{anchor points}, which are nearest with their corresponding prototypes. Then, the subset is used to pre-train a less-biased inference model to estimate the learning difficulties of different classes, which can be further utilized as the class-level sampling probabilities. Finally, a curriculum sampling module is designed to dynamically query most suitable samples from the rest data with the uncertainty estimation. In this way, the model is trained via the easy examples first and the hard samples later, preventing the unsuitable guidance at the early training stage~\cite{liu2021competence}.

Overall, our contributions can be summarized as follows: (1) We advocate that the training based on the \textit{learning difficulty} can filter the sub-optimal guidance at the early stage, which is crucial for the long-tailed skin lesion classification task. (2) Via considering the varied class distribution, the proposed flexible sampling framework achieves dynamic adjustments of sampling probabilities according to the class-level learning status, which leads to better performance. (3) Extensive experimental results demonstrate that our proposed model achieves new state-of-the-art long-tailed classification performance and outperforms other public methods on the ISIC dataset.

\section{Methodology}
\begin{figure}[t]
    \centering
    \includegraphics[width=12.2cm]{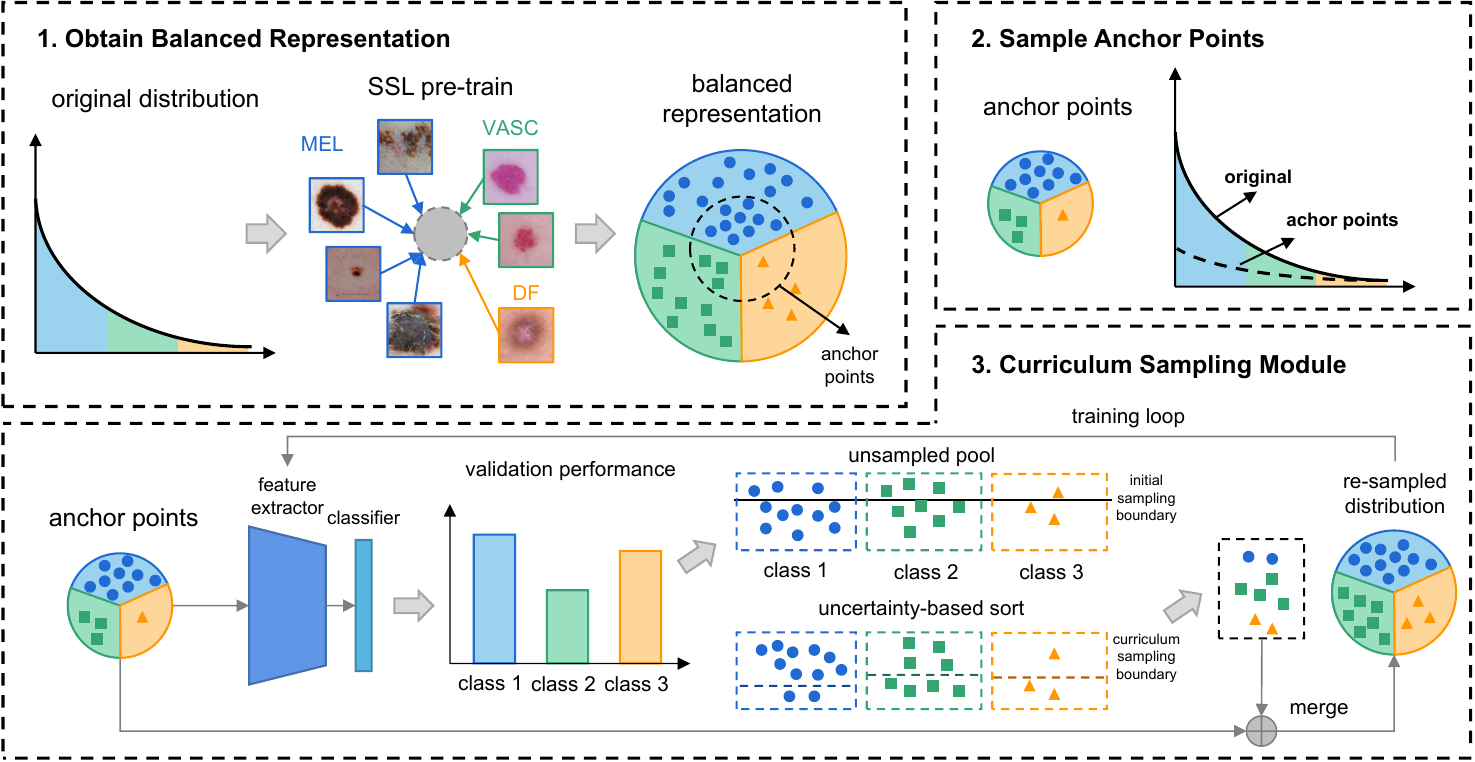}
    \caption{The overview of our proposed framework.}
    \label{fig.framework}
\end{figure}

The overall pipeline of our proposed flexible sampling framework is shown in Fig.~\ref{fig.framework}. First, we use self-supervised learning techniques to train a model to obtain balanced representation from original long-tailed distribution. Then, we sample a subset from the training data using the prototypes (mean class features) as anchor points. Next, this subset is used to train an inference model for the estimation of learning difficulties. Finally, a novel curriculum sampling module is proposed to dynamically query new instances from the unsampled pool according to the current learning status. An uncertainty-based selection strategy is used to help find the most suitable instances to query. 

\subsection{SSL Pre-training for Balanced Representations}
Training from the semantic labels as supervision signal on long-tailed distribution can bring significant bias since majority classes can occupy the dominant portion of the feature space~\cite{kang2019decoupling}. A straightforward solution is to leverage self-supervised learning (SSL) technique to obtain category distribution-agnostic features. 
Different from the supervised learning methods, SSL techniques do not require annotations but learn representations via maximizing the instance-wise discriminativeness~\cite{khosla2020supervised,kang2020exploring}. The contrastive learning loss~\cite{van2018representation} adopted in this study can be formulated as: 
\begin{equation}
    L_{CL} = \frac{1}{N}\sum_{i=1}^{N}-log\frac{exp(v_{i}\cdot\frac{v_{i}^{+}}{T})}{exp(v_{i}\cdot\frac{v_{i}^{+}}{T}) + \Sigma_{v_{i}^{-}\in V^{-}}exp(v_{i}\cdot\frac{v_{i}^{-}}{T})},
\end{equation}
where $v_{i}^{+}$ is a positive sample for $x_{i}$ after the data augmentation, $v^{-}$ is a negative sample drawn from any other samples, $T$ is the temperature factor. SSL can learn stably well feature spaces which is robust to the underlying distribution of a dataset, especially in a strongly-biased long-tailed distribution~\cite{kang2020exploring}.

\subsection{Sample Anchor Points using the Prototype}
To sample anchor points for training the inference model, we first comupte the mean features for each class as the prototype. Concretely, after pre-training a CNN model $M_{pre}$ with feature representation layer $f(\cdot)$ and classifier $h(\cdot)$, we can obtain features $[f(x_{1}), ..., f(x_{i}), ..., f(x_{n})]$ for n instances from original distribution $[x_{1}, ..., x_{i}, ..., x_{n}]$ by forward feeding into $f(\cdot)$.
Thus, for sorted class index $[c_{0}, ..., c_{j}, ..., c_{k-1}]$ where $N_{c_{0}} > ... > N_{c_{k-1}}$, the prototype $m_{c_{j}}$ of class $c_{j}$ can be calculated by averaging all the features from the same class:
\begin{equation}
    m_{c_{j}} = \frac{\sum_{i=1, x_{i}\in X_{c_{j}}}^{N_{c_{j}}} f(x_{i})}{N_{c_{j}}},
\end{equation}
where $N_{c{j}}$ denotes the number of instances of class $j$. Then, we find those representative samples which have the lowest distances from prototypes by calculating their Euclidean Distance:
\begin{equation}
    ds_{i, x_{i}\in c_{j}} = \left\|m_{c_{j}} - f(x_{i})\right\|_{2}.
\end{equation}
By sorting the distances $DS_{c_{k}} = [ds_{1}, ..., ds_{N_{c_{k}}}$], we can select those top-$\hat{N}_{c_{k}}$ lowest-distance samples as the class anchor points to form the subset. For instance, given a long-tailed distribution following Pareto Distribution with the number of classes $k$ and imbalance ratio $r = \frac{N_{0}}{N_{k-1}}$, the number of samples for class $c_{j} \in [0, k)$ can be calculated as $N_{c_{j}} = {(r^{-\frac{1}{(k-1)}})^{c_{j}}}*N_{0}$. Then, we use a scaling hyper-parameter $s \in (0, 1)$  as the imbalance ratio of anchor points $\hat{r} = r\cdot s$ and we can sample a less-imbalanced subset with $\hat{N}_{c_{j}} = {(\hat{r}^{-\frac{1}{(k-1)}})}^{c_{j}}*N_{0}$. 

Using the prototype for sampling anchor points show the following two advantages: 1) training the inference model with representative anchor points provides a better initialization compared to that of using random selected samples; 2) a re-sampled subset shows a more relatively balanced distribution, which helps reducing confirmation bias for predicting uncertainty on those unsampled instances.

\subsection{Curriculum Sampling Module}
\subsubsection{Class-wise Sampling Probability} After obtaining anchor points $\hat{X}$, we use them to train the inference model for querying new samples from the unsampled instances in the training pool $U_{X} = \complement_X\hat{X}$. As we claimed that there exist various learning difficulties and some tail classes with less intra-class variety can converge well in an early stage. Over-sampling such classes may bring obvious marginal effects and do harm to other classes. Hence, a more flexible sampling strategy is required with estimating the learning difficulty of each class. Intuitively, this can be achieved by evaluating the competence of the current model. Thus, we calculate the validation accuracy of different classes after every epoch as the sampling probability:
\begin{equation}
    p_{c_{j}} = 1-a_{e}(c_{j}),
\end{equation}
where $e$ denotes the $e_{th}$ epoch and $a_{e}(c_{j})$ is the corresponding validation accuracy on $j_{th}$ class. Then, the $p_{c_{j}}$ can be normalized as $\hat{p}_{c_{j}} = \delta \cdot p_{c_{j}}$, where $\delta = K/\sum_{j=0}^{K-1} p_{c_{j}}$. It should be noticed that when there are limited samples for validation, the $a_{e}(c_{j})$ can also be replaced by training accuracy with only one-time forward pass. The results evaluated on validation set can provide more unbiased reference with less risk of over-fitting since it exhibits different distribution from training set.
\subsubsection{Instance-wise Sample Selection} With learning difficulty-aware sampling probability at the categorical level, we need to determine which instances should be sampled for more efficient querying in an instance difficulty-aware manner. Here, we proposed to leverage uncertainty for the selection of unlabeled samples. Those samples with higher uncertainty have larger entropy since CNN model always give less-confident predictions on hard samples~\cite{leipnik1959entropy}. Training from high-uncertainty samples can well rich the representation space with discriminative features. Here, we compute the mutual information between predictions and model posterior for uncertainty measurement~\cite{houlsby2011bayesian}:
\begin{equation}
    \mathbb{I}(y_{i};\omega |x_{i}, X) = \mathbb{H}(y_{i}|x_{i}, X)-\mathbb{E}_{p(\omega|X)}[\mathbb{H}(y_{i}|x_{i}, X)].
\end{equation}
where $\mathbb{H}$ is the entropy of given instance point and $\mathbb{E}$ denotes the expectation of the entropy of the model prediction over the posterior of the model parameter. A high uncertainty value implies that the posterior draws are disagreeing among themselves~\cite{kirsch2019batchbald}. Then, with sorted instances $x_{1}, ..., x_{N_{c_{j}}}$ for class $j$, the number of newly sampling instances could be $\hat{p_{c_{j}}}\cdot N_{c_{j}}$. Finally, we merge the anchor points and newly-sampled instances into the training loop for a continual training.

\section{Experiments}
\subsection{Datasets}
We evaluate the competitive baselines and our methods on two dermatology datasets from ISIC~\cite{isic} with different number of classes. \textit{ISIC-2019-LT} is a long-tailed version constructed from ISIC-2019 Challenge\footnote{\url{https://challenge2019.isic-archive.com/}}, which aims to classify \textbf{8} kinds of diagnostic categories. We follow~\cite{liu2019large} and sample a subset from a Pareto distribution. With $k$ classes and imbalance ratio $r = \frac{N_{0}}{N_{k-1}}$, the number of samples for class $c_{j} \in [0, k)$ can be calculated as $N_{c_{j}} = {(r^{-\frac{1}{(k-1)}})^{c_{j}}}*N_{0}$. For this setting, we set $r = \{100, 200, 500\}$. We select 50 and 100 images from the remained samples as validation set and test set.
The second dataset is augmented from ISIC-2019 to increase the length of tail with extra 6
classes (\textbf{14} in total) added from the ISIC
Archive\footnote{\url{https://www.isic-archive.com/}} and named as
\textit{ISIC-Archive-LT}\footnote{{\color{magenta}{Please see our supplementary document for more details.}}}. This dataset is more challenging with longer length of tail and larger imbalance ratio $r \approx 1000$. The $ISIC-Archive-LT$ dataset is divided into train: val: test = 7:1:2.
\subsection{Implementation Details}
All images were resized into 224$\times$224 pixels. We used ResNet-18 and ResNet-50~\cite{he2016deep} as our backbones for \textit{ISIC-2019-LT} and \textit{ISIC-Archive-LT} respectively. For the SSL pre-training, we used Adam optimizer with a learning rate of 2$\times10^{-4}$ with a batch size of 16. For the classification task, we used Adam optimizer with a learning rate of 3$\times10^{-4}$ and a batch size of 256. Some augmentations techniques were applied such as random crop, flip and colour jitter. An early stopping monitor with the patience of 20 was set when there was no further increase on the validation set with a total of 100 epochs.

For our proposed methods, we warmed up the model with plain cross-entropy loss for 30 epochs. The scaling value $s$ for sampling anchor points was set as 0.1. We queried 10\% of the samples in the unsampled pool when there was no further improvement with the patience of 10. Then, the optimizer will be re-initialized with a learning rate of 3$\times10^{-4}$. For a fair comparison study, we kept all basic hyper-parameters the same on all the comparative methods. We reported the average of 5-trial running in all presented results. All experiments are implemented by PyTorch 1.8.1 platform on a single NVIDIA RTX 3090.

\subsection{Comparison Study}
The comparative methods selected in this study can be briefly introduced as follow: (1) plain class-balancing re-sampling (RS) and re-weighting (RW) strategies; (2) information augmentation: MixUp~\cite{zhang2017mixup} and OLTR~\cite{liu2019large}; (3) Modified re-weighting loss functions: Focal Loss~\cite{lin2017focal}, Class-Balancing (CB) Loss ~\cite{cui2019class} and label-distribution-aware margin (LDAM) Loss~\cite{cao2019learning}; (4) Curriculum learning-based methods: Curriculum Net~\cite{guo2018curriculumnet} and EarLy-exiting Framework (ELF)~\cite{duggal2020elf}. 
\subsubsection{Performance on \textit{ISIC-2019-LT}}
The overall results of the comparative methods and our proposed framework on~\textit{ISIC-2019-LT} are summarized in Table~\ref{table_isic-8} using the mean and standard deviation (Std.) in terms of top-1 accuracy. Here are some findings: (1) Re-sampling strategy obtained little improvements under an extremely imbalanced condition, e.g., r = 500. Repeatedly sampling those relatively easy-to-learn classes brings obvious marginal effects. (2) CB Loss showed superior performance with 4.56\%, 4.35\% and 5.05\% improvements on the baseline trained from cross-entropy, since it tended to sample more effective samples. (3) The proposed flexible sampling effectively outperformed other methods with 63.85\%, 59.38\% and 50.99\% top-1 accuracy over three different imbalance ratios.

\begin{table}[t]
\centering
\caption{The comparative results on three different imbalance ratios \{100, 200, 500\}.} 
\label{table_isic-8}
\setlength{\tabcolsep}{1.3mm}{\begin{tabular}{l|c|c|c}
\hline
\multicolumn{4}{c}{\textit{ISIC-2019-LT}} \\ \hline
\hline
\multicolumn{1}{l|}{Methods} & Top-1 @ Ratio 100 & Top-1 @ Ratio 200 & Top-1 @ Ratio 500 \\ \hline
CE                                               & 55.09 ($\pm$4.10)\scriptsize{\color{gray}{=0.00}}       &  53.40 ($\pm$0.82)\scriptsize{\color{gray}{=0.00}}     &   44.40 ($\pm$1.80)\scriptsize{\color{gray}{=0.00}}                \\
RS                                               & 61.05 ($\pm$1.26)\scriptsize{\color{ao}{+4.96}}       &  55.10 ($\pm$1.29)\scriptsize{\color{ao}{+1.70}}     &  46.82 ($\pm$1.33)\scriptsize{\color{ao}{+2.82}}                 \\
RW                                               & 61.07 ($\pm$1.73)\scriptsize{\color{ao}{+5.98}}       &  58.03 ($\pm$1.24)\scriptsize{\color{ao}{+4.63}}     &  50.55 ($\pm$1.19)\scriptsize{\color{ao}{+6.15}}                 \\ \hline
MixUp~\cite{zhang2017mixup}                      & 57.25 ($\pm$0.71)\scriptsize{\color{ao}{+2.16}}       &  52.05 ($\pm$1.71)\scriptsize{\color{red}{$-$1.35}}     & 42.78 ($\pm$1.28)\scriptsize{\color{red}{$-$1.62}}                  \\
OLTR~\cite{liu2019large}                         & 60.95 ($\pm$2.09)\scriptsize{\color{ao}{+5.86}}       &  56.77 ($\pm$1.99)\scriptsize{\color{ao}{+3.37}}     &  48.75 ($\pm$1.52)\scriptsize{\color{ao}{+4.35}}                 \\ \hline
Focal Loss~\cite{lin2017focal}                   & 60.62 ($\pm$1.04)\scriptsize{\color{ao}{+5.53}}       &  53.88 ($\pm$1.19)\scriptsize{\color{ao}{+0.48}}     &  44.38 ($\pm$1.44)\scriptsize{\color{red}{$-$0.02}}                 \\
CB Loss~\cite{cui2019class}                      & 59.65 ($\pm$1.52)\scriptsize{\color{ao}{+4.56}}       &  57.75 ($\pm$1.39)\scriptsize{\color{ao}{+4.35}}     &  49.45 ($\pm$1.11)\scriptsize{\color{ao}{+5.05}}                 \\
LDAM~\cite{cao2019learning}                      & 58.62 ($\pm$1.46)\scriptsize{\color{ao}{+3.53}}       &  54.92 ($\pm$1.24)\scriptsize{\color{ao}{+1.52}}     &  44.90 ($\pm$2.60)\scriptsize{\color{ao}{+0.50}}                 \\ \hline
CN~\cite{guo2018curriculumnet}        & 56.41 ($\pm$1.62)\scriptsize{\color{ao}{+1.32}}       &  53.55 ($\pm$0.96)\scriptsize{\color{ao}{+0.15}}     &  44.92 ($\pm$0.31)\scriptsize{\color{ao}{+0.52}}                 \\
ELF~\cite{duggal2020elf}                         & 61.94 ($\pm$1.53)\scriptsize{\color{ao}{+6.85}}       &  54.60 ($\pm$1.78)\scriptsize{\color{ao}{+1.20}}     &  43.19 ($\pm$0.31)\scriptsize{\color{red}{$-$1.21}}                 \\ \hline 
Ours                                             &  \textbf{63.85 ($\pm$2.10)\scriptsize{\color{ao}{+8.76}}}       & \textbf{59.38 ($\pm$1.90)\scriptsize{\color{ao}{+5.98}}}       &  \textbf{50.99 ($\pm$2.02)\scriptsize{\color{ao}{+6.59}}}                 \\ \hline
\end{tabular}}
\end{table}

\begin{table}[t]
\caption{The comparative results on \textit{ISIC-Archive-LT}.} 
\label{table_isic-arxiv}
\centering
\setlength{\tabcolsep}{2mm}{\begin{tabular}{lcccc}
\hline
\multicolumn{5}{c}{\textit{ISIC-Archive-LT}} \\ \hline
\hline
\multicolumn{1}{l|}{Methods} & \multicolumn{1}{c|}{Head}  & \multicolumn{1}{c|}{Medium} & \multicolumn{1}{c|}{Tail}   & All   \\ \hline
\multicolumn{1}{l|}{CE}      & \multicolumn{1}{c|}{71.23 ($\pm$1.33)} & \multicolumn{1}{c|}{48.73 ($\pm$4.24)}  & \multicolumn{1}{c|}{36.94 ($\pm$7.17) } & 52.30 ($\pm$4.25)  \\
\multicolumn{1}{l|}{RS}      & \multicolumn{1}{c|}{70.26  ($\pm$1.21)} & \multicolumn{1}{c|}{55.98 ($\pm$1.63)} & \multicolumn{1}{c|}{37.14 ($\pm$5.44)} & 54.46 ($\pm$2.76)\\
\multicolumn{1}{l|}{RW}      & \multicolumn{1}{c|}{59.70 ($\pm$2.84)} & \multicolumn{1}{c|}{61.53 ($\pm$3.69)}  & \multicolumn{1}{c|}{63.64 ($\pm$3.11)} & 61.62 ($\pm$3.21)\\
\multicolumn{1}{l|}{CBLoss}  & \multicolumn{1}{c|}{65.27 ($\pm$5.73)} & \multicolumn{1}{c|}{56.97 ($\pm$6.64)}  & \multicolumn{1}{c|}{66.74 ($\pm$2.94)}& 62.89 ($\pm$5.10)\\
\multicolumn{1}{l|}{ELF}     & \multicolumn{1}{c|}{68.81 ($\pm$3.69)} & \multicolumn{1}{c|}{50.17 ($\pm$4.12)} & \multicolumn{1}{c|}{60.77 ($\pm$4.99)} & 59.92 ($\pm$4.27) \\ \hline
\multicolumn{1}{l|}{Ours}    & \multicolumn{1}{c|}{68.26 ($\pm$2.01)} & \multicolumn{1}{c|}{57.72 ($\pm$2.98)} & \multicolumn{1}{c|}{67.01 ($\pm$4.12)} & \textbf{64.33 ($\pm$3.04)} \\ \hline
\end{tabular}}
\end{table}

\subsubsection{Performance on \textit{ISIC-Archive-LT}}
\begin{figure}[t]
    \centering
\includegraphics[width=9cm]{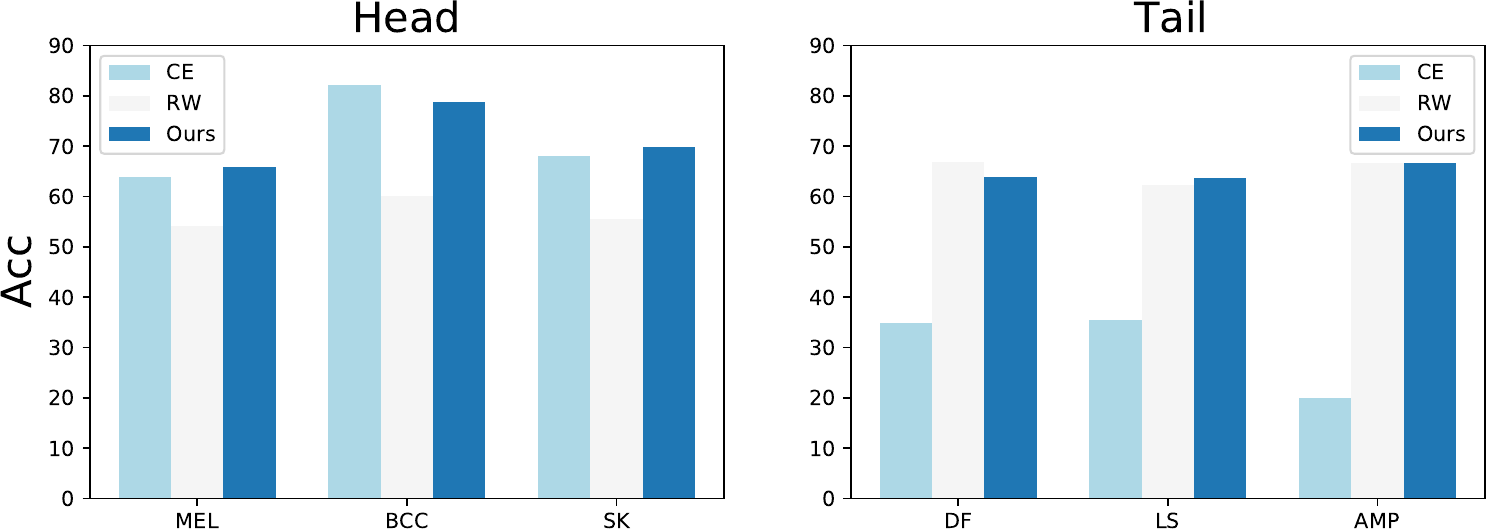}
    \caption{The performance on several classes from head/tail classes on three methods.}
    \label{fig.head-tail}
\end{figure}

For \textit{ISIC-Archive-LT} dataset with imbalanced test set, we reported the accuracy on shot-based division, e.g., head, medium, tail and their average results as most existing works~\cite{liu2019large,zhang2021deep}.
The overall results are shown in Table~\ref{table_isic-arxiv}. RS did not obtain satisfactory results. It is due to that the same training samples from tail classes can be occurred repeatedly in a mini-batch during the iteration under an extremely imbalanced condition, resulting in learning from an incomplete distribution. Compared with CBLoss which sacrificed head classes performance, our methods evenly improved the overall performance (+12.03\%) with the medium (+8.99\%) and tail classes (+30.07\%) but only a little performance loss on head classes (-2.97\%). In Fig.~\ref{fig.head-tail} we present the performance of several selected classes evaluated on three competitors. It is noticed that flex sampling can also improve those head classes with more learning difficulty, e.g., MEL, while RW severely hurts the recognition performance especially for those high-accuracy classes, e.g., BCC.

\subsection{Ablation Study} 
To investigate what makes our proposed methods performant, we conduct ablation studies and report the results in Table~\ref{table_ablation}. We first evaluate the selection of pre-training methods. It can be found that using SSL pre-training with anchor points can well boost the accuracy of the inference model, where \textit{Random} denotes the random selection of samples and \textit{Edge Points} denotes those samples away from the prototypes. Finally, we validated the effect of different querying strategies. Compared with random querying, uncertainty-based mutual information can help introduce samples more efficiently in an instance-aware manner.

\begin{table}[t]
\caption{Ablation study results on \textit{ISIC-2019-LT} ($r$ = 100).}
\label{table_ablation}
\centering
\begin{tabular}{c|c|clcc}
\cline{1-3} \cline{5-6}
\multicolumn{1}{l|}{Pre-train} & Sample Selection & Acc.              &  & \multicolumn{2}{c}{+Querying Strategy} \\ \cline{1-3} \cline{5-6} 
CE                             & Anchor Points    & 58.09 ($\pm$0.95) &  & Random             & Mutual Information              \\ \cline{1-3} \cline{5-6} 
\multirow{3}{*}{SSL}           & Random           & 56.19 ($\pm$2.20) &  & 57.77 ($\pm$3.09)             & 60.99 ($\pm$2.41)             \\
                               & Edge Points      & 52.62 ($\pm$1.76) &  & 56.95 ($\pm$1.04)             & 58.63 ($\pm$1.07)            \\
                               & Anchor Points    & \textbf{59.32 ($\pm$1.58)} &  & 60.26 ($\pm$1.73)             & \textbf{63.85 ($\pm$2.10)}             \\ \cline{1-3} \cline{5-6} 
\end{tabular}
\end{table}

\section{Conclusion}
In this work, we present a flexible sampling framework with taking into consideration the learning difficulties of different classes. We advocate training from a relatively balanced and representative subset and dynamically adjusting the sampling probability according to the learning status. The comprehensive experiments demonstrate the effectiveness of our proposed methods. In future work, more medical datasets and a multi-label setting can be further explored.

\bibliographystyle{splncs04}
\bibliography{ref}

\begin{thebibliography}{10}
\providecommand{\url}[1]{\texttt{#1}}
\providecommand{\urlprefix}{URL }
\providecommand{\doi}[1]{https://doi.org/#1}

\bibitem{cao2019learning}
Cao, K., Wei, C., Gaidon, A., Arechiga, N., Ma, T.: Learning imbalanced
  datasets with label-distribution-aware margin loss. Advances in neural
  information processing systems  \textbf{32} (2019)

\bibitem{cui2019class}
Cui, Y., Jia, M., Lin, T.Y., Song, Y., Belongie, S.: Class-balanced loss based
  on effective number of samples. In: Proceedings of the IEEE/CVF conference on
  computer vision and pattern recognition. pp. 9268--9277 (2019)

\bibitem{duggal2020elf}
Duggal, R., Freitas, S., Dhamnani, S., Chau, D.H., Sun, J.: Elf: An
  early-exiting framework for long-tailed classification. arXiv preprint
  arXiv:2006.11979  (2020)

\bibitem{england2016diagnostic}
England, N., Improvement, N.: Diagnostic imaging dataset statistical release.
  London: Department of Health  \textbf{421} (2016)

\bibitem{esteva2017dermatologist}
Esteva, A., Kuprel, B., Novoa, R.A., Ko, J., Swetter, S.M., Blau, H.M., Thrun,
  S.: Dermatologist-level classification of skin cancer with deep neural
  networks. nature  \textbf{542}(7639),  115--118 (2017)

\bibitem{guo2018curriculumnet}
Guo, S., Huang, W., Zhang, H., Zhuang, C., Dong, D., Scott, M.R., Huang, D.:
  Curriculumnet: Weakly supervised learning from large-scale web images. In:
  Proceedings of the European Conference on Computer Vision (ECCV). pp.
  135--150 (2018)

\bibitem{he2016deep}
He, K., Zhang, X., Ren, S., Sun, J.: Deep residual learning for image
  recognition. In: Proceedings of the IEEE conference on computer vision and
  pattern recognition. pp. 770--778 (2016)

\bibitem{houlsby2011bayesian}
Houlsby, N., Husz{\'a}r, F., Ghahramani, Z., Lengyel, M.: Bayesian active
  learning for classification and preference learning. arXiv preprint
  arXiv:1112.5745  (2011)

\bibitem{isic}
ISIC: Isic archive (2021), \url{https://www.isic-archive.com/}

\bibitem{ju2021relational}
Ju, L., Wang, X., Wang, L., Liu, T., Zhao, X., Drummond, T., Mahapatra, D., Ge,
  Z.: Relational subsets knowledge distillation for long-tailed retinal
  diseases recognition. In: Medical Image Computing and Computer Assisted
  Intervention -- MICCAI 2021. pp. 3--12. Springer, Cham (2021)

\bibitem{kang2020exploring}
Kang, B., Li, Y., Xie, S., Yuan, Z., Feng, J.: Exploring balanced feature
  spaces for representation learning. In: International Conference on Learning
  Representations (2020)

\bibitem{kang2019decoupling}
Kang, B., Xie, S., Rohrbach, M., Yan, Z., Gordo, A., Feng, J., Kalantidis, Y.:
  Decoupling representation and classifier for long-tailed recognition. arXiv
  preprint arXiv:1910.09217  (2019)

\bibitem{khosla2020supervised}
Khosla, P., Teterwak, P., Wang, C., Sarna, A., Tian, Y., Isola, P., Maschinot,
  A., Liu, C., Krishnan, D.: Supervised contrastive learning. Advances in
  Neural Information Processing Systems  \textbf{33},  18661--18673 (2020)

\bibitem{kirsch2019batchbald}
Kirsch, A., Van~Amersfoort, J., Gal, Y.: Batchbald: Efficient and diverse batch
  acquisition for deep bayesian active learning. Advances in neural information
  processing systems  \textbf{32} (2019)

\bibitem{leipnik1959entropy}
Leipnik, R.: Entropy and the uncertainty principle. Information and Control
  \textbf{2}(1),  64--79 (1959)

\bibitem{lin2017focal}
Lin, T.Y., Goyal, P., Girshick, R., He, K., Doll{\'a}r, P.: Focal loss for
  dense object detection. In: Proceedings of the IEEE international conference
  on computer vision. pp. 2980--2988 (2017)

\bibitem{liu2021competence}
Liu, F., Ge, S., Wu, X.: Competence-based multimodal curriculum learning for
  medical report generation. In: Proceedings of the 59th Annual Meeting of the
  Association for Computational Linguistics and the 11th International Joint
  Conference on Natural Language Processing (Volume 1: Long Papers). pp.
  3001--3012 (2021)

\bibitem{liu2019large}
Liu, Z., Miao, Z., Zhan, X., Wang, J., Gong, B., Yu, S.X.: Large-scale
  long-tailed recognition in an open world. In: Proceedings of the IEEE/CVF
  Conference on Computer Vision and Pattern Recognition. pp. 2537--2546 (2019)

\bibitem{van2018representation}
Van~den Oord, A., Li, Y., Vinyals, O.: Representation learning with contrastive
  predictive coding. arXiv e-prints pp. arXiv--1807 (2018)

\bibitem{wang2020long}
Wang, X., Lian, L., Miao, Z., Liu, Z., Yu, S.X.: Long-tailed recognition by
  routing diverse distribution-aware experts. arXiv preprint arXiv:2010.01809
  (2020)

\bibitem{zhang2017mixup}
Zhang, H., Cisse, M., Dauphin, Y.N., Lopez-Paz, D.: mixup: Beyond empirical
  risk minimization. arXiv preprint arXiv:1710.09412  (2017)

\bibitem{zhang2021deep}
Zhang, Y., Kang, B., Hooi, B., Yan, S., Feng, J.: Deep long-tailed learning: A
  survey. arXiv preprint arXiv:2110.04596  (2021)

\bibitem{zhou2020bbn}
Zhou, B., Cui, Q., Wei, X.S., Chen, Z.M.: Bbn: Bilateral-branch network with
  cumulative learning for long-tailed visual recognition. In: Proceedings of
  the IEEE/CVF Conference on Computer Vision and Pattern Recognition. pp.
  9719--9728 (2020)

\end{thebibliography}
\end{document}


%
\title{Supplementary Document}
\titlerunning{Flexible Sampling}
\author{Paper ID: 166}
\authorrunning{PID: 166}
\institute{***}

\maketitle              

\begin{figure}
    \centering
    \includegraphics[width=12cm]{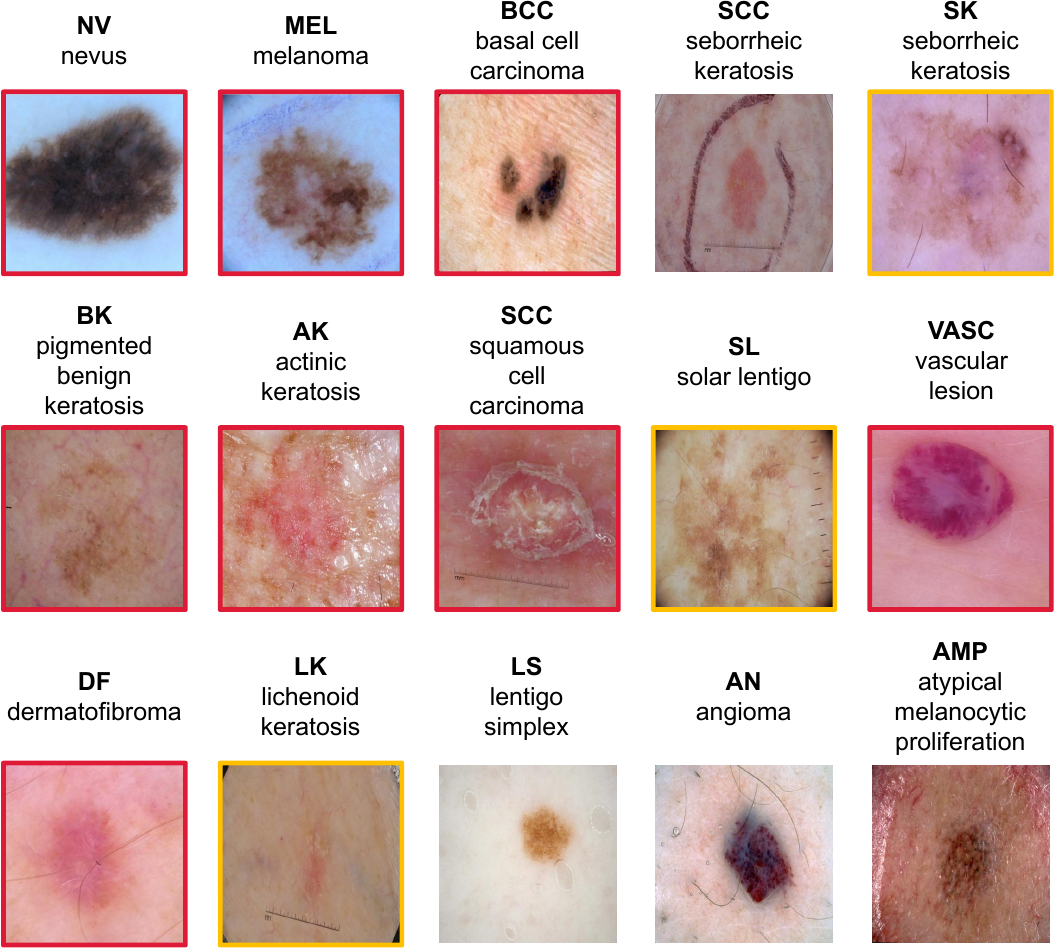}
    \caption{An illustration of 15 classes in our study. Those classes with \color{red}{Red bounding box} \color{black} constitute \textit{ISIC-2019-LT}. Then, in the second dataset \textit{ISIC-Archive-LT}, BK is further divided into three sub-types: SK, SL and LK, which are denoted by \color{orange}{Orange bounding box}. \color{black}{}There are \textbf{8} and \textbf{14} classes in two datasets respectively.}
    \label{fig:my_label}
\end{figure}

\begin{table}[t]
\centering
\caption{The samples of each class in \textit{ISIC-Archive-LT} and the group division. }
\begin{tabular}{cccccccccccccccc}
\hline
\hline

\multicolumn{15}{c}{ISIC-Archive-LT}                                                                                                                             \\ \hline
\multicolumn{1}{c|}{Group} & \multicolumn{4}{c|}{Many}                      & \multicolumn{4}{c|}{Medium}                      & \multicolumn{6}{c}{Few}         \\ \hline
\multicolumn{1}{c|}{Abb.}  & NV   & MEL  & BCC  & \multicolumn{1}{c|}{SK}    & AK  & SCC & BKL & \multicolumn{1}{c|}{SL}  & VASC & DF  & LK & LS & AN & AMP \\ \hline
\multicolumn{1}{c|}{Train} & 9012 & 3165 & 2375 & \multicolumn{1}{c|}{1024}  & 608 & 459 & 268 & \multicolumn{1}{c|}{189} & 177  & 172 & 22 & 18 & 10 & 9   \\
\multicolumn{1}{c|}{Val}   & 1288 & 452  & 339  & \multicolumn{1}{c|}{147}   & 87  & 65  & 39  & \multicolumn{1}{c|}{27}  & 25   & 24  & 3  & 3  & 2  & 2   \\
\multicolumn{1}{c|}{Test}  & 2575 & 905  & 679  & \multicolumn{1}{c|}{293}   & 174 & 132 & 77  & \multicolumn{1}{c|}{54}  & 51   & 50  & 7  & 6  & 3  & 3   \\ \hline \hline
\end{tabular}
\end{table}

\newpage

\begin{python}[t]
# The implementation of BALD in a pytorch-like programming
probs = torch.zeros([n_drop, len(data), len(np.unique(Y))])
for i in range(n_drop):
    for x, y, idxs in loader:
        out, e1 = model(x)
        prob = F.softmax(out, dim=1)
        probs[i][idxs] += F.softmax(out, dim=1)
pb = probs.mean(0)
entropy1 = (-pb*torch.log(pb)).sum(1)
entropy2 = (-probs*torch.log(probs)).sum(2).mean(0)
uncertainties = entropy2 - entropy1

\end{python}